\documentclass{article}

\usepackage{arxiv}

\usepackage[utf8]{inputenc} 
\usepackage[T1]{fontenc}    
\usepackage{hyperref}       
\usepackage{url}            
\usepackage{booktabs}       
\usepackage{amsfonts}       
\usepackage{nicefrac}       
\usepackage{microtype}      
\usepackage{lipsum}
\usepackage{graphicx}
\usepackage{indentfirst}
\usepackage{graphicx}
\usepackage{dcolumn}
\usepackage{bm}
\usepackage[inline]{enumitem}
\usepackage{placeins}
\usepackage{graphicx}
\usepackage{amsfonts}
\usepackage{xcolor}
\usepackage{tcolorbox}
\usepackage{filecontents,lipsum}
\usepackage{physics}
\usepackage{xcolor}
\usepackage{appendix}
\usepackage{fullpage}
\usepackage{multirow}
\usepackage{graphicx}
\usepackage{array}
\newcommand\numberthis{\addtocounter{equation}{1}\tag{\theequation}}

\title{Construction Of 'Support Vector' Machine Feature Spaces Via Deformed Weyl-Heisenberg Algebra}

\author{
  Shahram Dehdashti, \hspace{1cm } Catarina  Moreira, \hspace{1cm }  Abdul Karim Obeid,\\
   \AND
   Peter Bruza $^{1}$  \\
   School of Information Systems\\ Queensland University of Technology\\ Brisbane, Australia\\
  \texttt{p.bruza@qut.edu.au} }

\begin{document}

\maketitle

\begin{abstract}
This paper uses deformed coherent states, based on a deformed Weyl-Heisenberg algebra that unifies the well-known $SU(2)$, Weyl-Heisenberg, and $SU(1,1)$ groups, through a common  parameter $\alpha$.
We show that deformed coherent states provide the theoretical foundation of a meta-kernel function, that is a kernel which in turn defines kernel functions.
Kernel functions drive developments in the field of machine learning and the meta-kernel function presented in this paper opens new theoretical avenues for the definition and exploration of kernel functions.
The meta-kernel function applies associated revolution surfaces  as feature spaces identified with non-linear coherent states.  
An empirical investigation compares the $\alpha-SU(2)$ and $\alpha-SU(1,1)$ kernels derived from the meta-kernel which shows performance similar to the Radial Basis kernel, and offers new insights (based on the deformed Weyl-Heisenberg algebra).
\end{abstract}
\footnote{The first three authors contributed equally in producing the majority of the content. SD developed the theory. CM verified and analysed the empirical results and conducted the grid search. AO conducted produced the initial computational model of the theory with initial empirical results. PB coordinated the research and revised the manuscript.}%


\section{Introduction}



    Support Vector Machines (SVMs) are one of the most widely used algorithms for classification problems. Originally proposed in the works of Boser et al~\cite{boser1992training} and Cortes and Vapnik~\cite{cortes1995support}, they can be defined as learning machines which construct an $n$-dimensional decision boundary surface (also called a hyperplane) that optimally separates data into positive and negative classes by maximizing the margin of separation between them. 
    
    Contrary to artificial neural networks, which provide a local minimum solution to the optimization problem, SVMs provide a unique globally optimal solution for the margin separation problem, which is addressed through the application of a kernel-based learning method. In this context, a kernel is understood as a similarity function that is applied to each data point to map the original non-linear observations into a higher dimensional space where the observations may become linearly separable. A wide range of different kernels have been proposed in the literature, targeting specific classification problems~\cite{Haykin08}. The Gaussian kernel (also referred to as the Radial Basis Function kernel (RBF)), is probably the most widely used kernel demonstrating 'state of the art' performance in a variety of  classification problems~\cite{Bhattacharyya11}.  \\
\indent More recently, new mappings between non-linear separable observations and higher dimensional feature spaces have been proposed with the purpose of extending the capabilities of SVMs towards theoretically feasible quantum-inspired machine learning algorithms~\cite{schuld2019quantum,schuld2019machine,havlivcek2019supervised,mehta2019high,killoran2019strawberry,killoran2018continuous,srinivasan2018learning,adhikary2019supervised,bartkiewicz2019experimental}.  For instance, under a quantum theoretical perspective, by mapping data into coherent states which are a superposition of eigen-functions of a quantum harmonic oscillator with minimum Heisenberg  uncertainty, the RBF kernel can be understood as the inner product of two coherent states  \cite{kubler2019quantum}. The coherent state of this harmonic oscillator comprises the following properties:
\begin{enumerate*}[label=({\it \roman*})]
  \item It is obtained by the displacement operators on the ground state;
  \item It is an eigenfunction of the annihilation operator;
  \item It satisfies the minimum uncertainty relation, i.e., $\Delta(\mathbf{x})=\Delta(\mathbf{p})=\sigma/\sqrt{2}$, in which $\Delta(\mathbf{x})$ and $\Delta(\mathbf{p})$ are respectively the variance of the position and momentum of the harmonic oscillator;
  \item It is over-complete.
\end{enumerate*}
\\
\indent 
The over-completeness property implies an arbitrary function can be expressible as a linear combination of kernel functions in a "reproducing Hilbert space" \cite{combescure2012coherent}. Any of the first three above-mentioned properties lead to a definition of generalized coherent states, although property ({\it iv}) is necessary for the definition of coherent states. For example, while a Gazeau-Klauder coherent state is defined by property ({\it ii}) and fulfils property ({\it iii}), displacement-type coherent states are obtained by displacement operators on reference states  \cite{ali2000coherent}.
\indent 
Recently, Schuld and Killoran proposed to map data from the original space to a feature space by using squeezed coherent states \cite{schuld2019quantum}. Squeezed  states are coherent states that saturate the Heisenberg uncertainty principle in such a way that the variance of position and momentum depend on a so-called squeezed parameter. Therefore, the reduced uncertainty is one of its quadrature components, while increased uncertainty is the latter, i.e., $\Delta(\mathbf{x})=\exp{\zeta}/\sqrt{2}$ and $\Delta(\mathbf{p})=\exp{-\zeta}/\sqrt{2}$, where $\zeta$ is the  squeezing parameter. The squeezing parameter controls uncertainty via a quadrature component, while the third  property of coherent states are preserved.\\
\indent 
Given the large number of kernel functions currently being proposed in the literature, the question naturally arises as to which kernel function to apply. In SVM-based classification problems, the appropriate choice of a kernel is fundamental, however, the current 'trial-and-error' nature of selecting the best kernel poses significant challenges, especially when one considers kernels that can support both classical and quantum-inspired machine learning algorithms, which renders the kernel choice problem an open research question \cite{Ali06}.\\
\indent To address this problem (and taking as basis the work of Schuld and Killoran on squeezed coherent states~\cite{schuld2019quantum}), we propose a generalised meta-kernel from which the RBF kernel (and other kernels) can be derived by using a deformed Weyl-Heisenberg (dW-H) algebra, dependent on a parameter $\alpha \in \mathbb{R}$. By applying the associated displacement operator on the reference state, the non-linear coherent state is generated by considering the specific value of the parameter, i.e. $\alpha=2$, $\alpha=0$, and $\alpha=-2$, $SU(2)$, W-H, and $SU(1,1)$-coherent state are respectively recovered. \cite{dehdashti2013coherent,dehdashti2015realization,castillo2019polynomial}.  
The choice of $\alpha$ allows a specific kernel function to be defined. Therefore, the theory of coherent states can be seen as providing a meta-kernel from which kernel functions can be derived. To the best of our knowledge, no such theory of meta-kernels presently exists.
 
\begin{figure}[t]
  \centering
  \includegraphics[scale=.55]{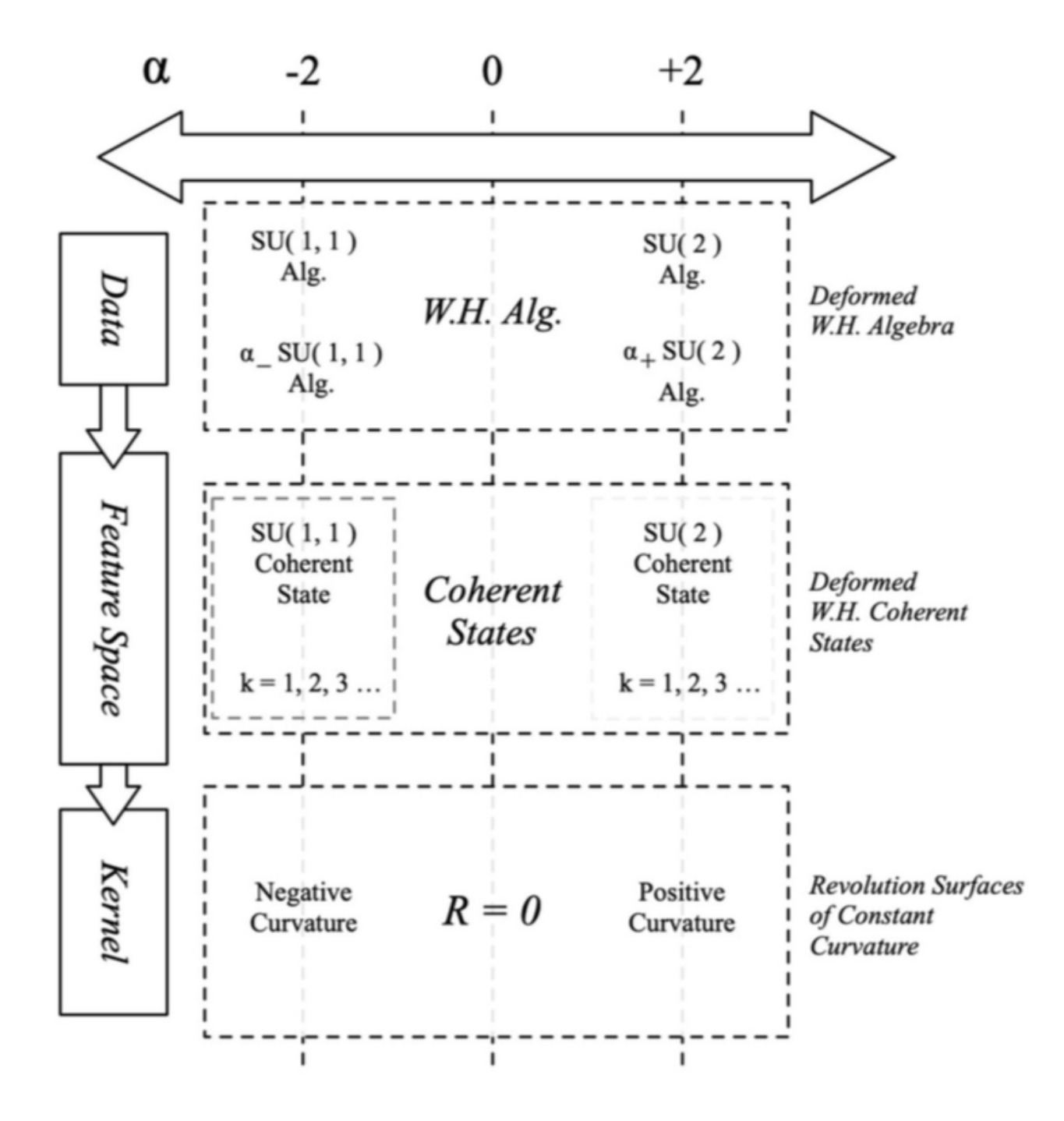}
  \caption{Schematic representation of the SVM method based on the Weyl-Heisenberg algebra, showing the mapping of data into the feature space, represented by the dW-H coherent state.}
  \label{fig1}
\end{figure}

 By means of a feature mapping, data is mapped into the feature space, represented by the deformed coherent space. Schematically, this is illustrated in Figure \ref{fig1}. Geometrically, the feature space constructed by the dW-H coherent state is a surface of revolution with constant curvature, i.e., the surfaces associated with $\alpha-SU(2)$ and $\alpha-SU(1,1)$ are respectively a positive compact surface, and negative surface, while $\alpha=0$ produces a flat surface. Therefore, a kernel function defined in any one of the configurations is an inner product of two elements on the related surface. Through this process, our dW-H algebra acts like a meta-theory from which  a new class of two parameter non-linear kernel functions can be derived.


The paper will proceed as follows. In Section \ref{coherent_states}, we provide a brief introduction to dW-H coherent states which are then expressed in kernel functions. We also describe the geometric properties of the feature spaces in which these kernel functions are defined. In Section \ref{test_design}, a test design is formulated for an illustrative evaluation of the introduced kernel functions, from the standpoint of enrichment of Gaussian strategies in SVM classification. 
Accompanying empirical results are presented, along with visualisations that aid descriptions of relevant observations.  Section \ref{discussion} discusses the benefits of the algebra within SVM classification based on the results of the empirical evaluation.
\section{Deriving kernel functions from deformed Coherent States}
\label{coherent_states}
A supervised machine learning (ML) classification problem can be formalised in the following way. Given a set of $N$ training examples $\{ (x_{1}, y_{1}), (x_{2}, y_{2}), \cdots, (x_{N}, y_{N})\}$, where $x_{i}$ corresponds to the $ith$ training example, each training example is represented by a set of input features, and $y_{i}$ corresponds to the 'ground truth' label of the training example $x_{i}$, the objective of ML is to learn a model, $h(x)$, that represents the training set. Ideally, the outcome is to generate the model that is most capable of correctly predicting the class labels of unseen instances.

One way of predicting unseen examples is through the application of a similarity function, a \textit{kernel}, between the unseen input instance ${x^{\prime}}$ and each of the training inputs, ${x_i}$, learned during the training phase.

Kernel methods, $K(x, x^{\prime})$, use the inner product between any two inputs $x, x^{\prime} \in \mathcal{X}$, as distance measures in order to construct models that capture the properties of a data distribution. These distance measures can be defined in a feature space $\mathcal{X}$, depending on whether the data is linear, or non-linearly separable.

\begin{figure}[t]
  \centering
  \includegraphics[scale=0.6]{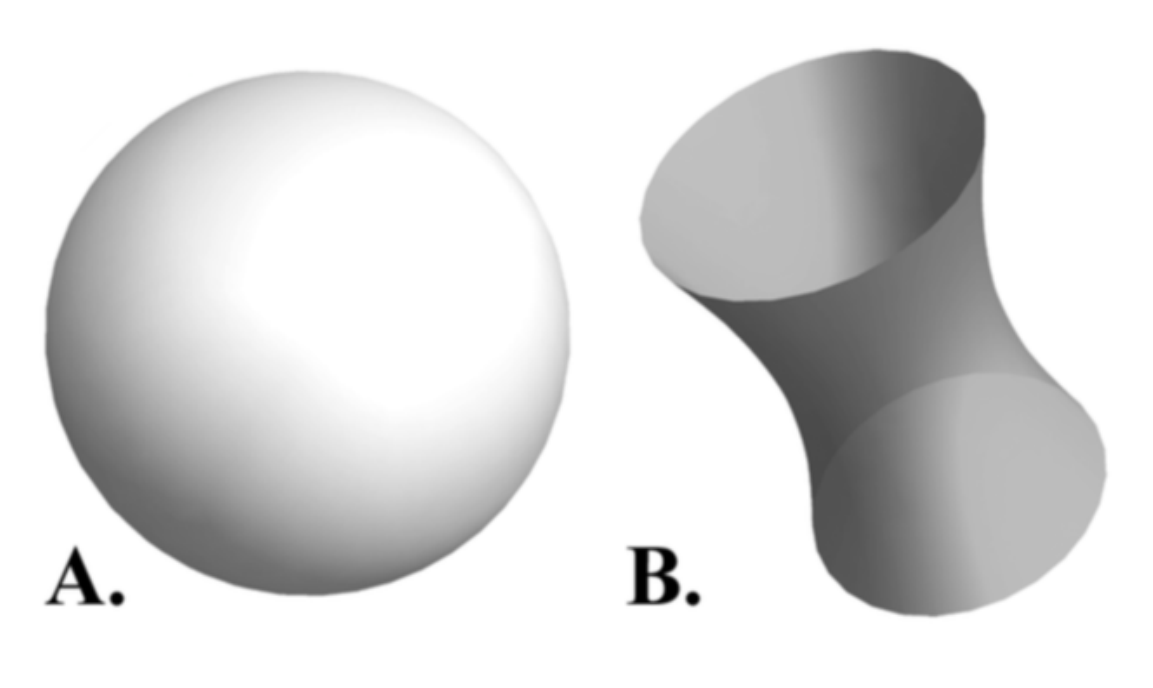}
  \caption{Spatial definition of feature spaces for kernel functions $\alpha-SU(2)$ \textbf{(A.)} And $\alpha-SU(1,1)$ \textbf{(B.)} Under parametric configuration $\alpha = 2$ And $k = 1$}
  \label{parametric_feature_spaces}
\end{figure}

The left side of Figure \ref{fig1} schematically represents this process. One can define a complex Hilbert space as the feature space, where a feature mapping $\phi: \mathcal{X}\rightarrow \mathcal{F}$, in which $\mathcal{F}$ is a complex Hilbert state,   $\phi: x \rightarrow |\phi(x) \rangle $, implies a kernel function can be defined as $K(x, x^{\prime} ) =
 \langle\phi(x), \phi(x^{\prime}) \rangle$. By operating a dW-H algebra on the reference state, the feature space is constructed with deformed coherent states. These coherent states depend on the attributed sign of the parameter $\alpha$, meaning `positivity' or `negativity' defines an $\alpha-SU(2)$ coherent state or $\alpha-SU(1,1)$ coherent state respectively; while $\alpha=0$ stands for a harmonic oscillator coherent state ( see Figure \ref{fig1}). For the sake of simplicity, we define dW-H coherent states for positive and negative values, separately. The first is titled the $\alpha-SU(2)$ coherent state, defined as follows:
 \begin{eqnarray}\label{eq3}
|x;z,\alpha\rangle
&=&\left[1+\tan^{2}\left(z\sqrt{\frac{\alpha}{2}}\right)\right]^{-k}\sum_{m=0}^{2k} (-1)^{m} e^{-imx}\nonumber\\
&\times&\tan^{m}\left(z\sqrt{\frac{\alpha}{2}}\right) \sqrt{\frac{(2k)!}{m!(2k-m)!}}|k,m\rangle,
\end{eqnarray} 
 in which $\alpha \geq 0$, and  $k \in N$. The second is named the $\alpha-SU(1,1)$ coherent state:
 \begin{eqnarray}\label{eq2}
|x;z,\alpha\rangle&=&\left[1-\tanh^{2}\left(z\sqrt{\frac{|\alpha|}{2}}\right)\right]^{k}\sum_{m=0}^{\infty} (-1)^{m} e^{-imx}\nonumber\\
&\times&\tanh^{m}\left(z\sqrt{\frac{|\alpha|}{2}}\right) \sqrt{\frac{\Gamma[2k+m]}{m!\Gamma[2k]}}|k,k+m\rangle,
\end{eqnarray} 
where $\alpha\leq 0$, and $k\in N$. 
 Note that in the case of $\alpha = \pm 2$, coherent states (\ref{eq3}) and (\ref{eq2}), respectively reduce $SU(2)$ and $SU(1,1)$ coherent states. It was also shown that if $\alpha $ approaches zero, both coherent states (\ref{eq3}) and (\ref{eq2}) reduce to the harmonic oscillator coherent states  \cite{dehdashti2013coherent} , i.e.,
 \begin{eqnarray}
 |z\rangle=e^{-|z|^{2}}\sum_{m=0}^{\infty} \frac{z^{m} }{\sqrt{m!}}|m
 \rangle.
 \end{eqnarray}
 \indent Hence,  by considering a multi-dimensional input set in a data set of vectors $\mathbf{x} = (x_{1},\cdots,x_{N})^{T} \in \mathbb{R}^{N}$, one can define the joint state of $N$ deformed coherent states,
 \begin{align*}
 & \phi: (x_{1},\cdots,x_{N}) \rightarrow \\ & |x_{1},z,\alpha\rangle\otimes |x_{2},z,\alpha\rangle \otimes \cdots \otimes |x_{N},z,\alpha\rangle. \numberthis
 \end{align*}{}
 Therefore, the kernel  is defined as the following:
 \begin{eqnarray}
 K(\mathbf{x},\mathbf{x^{\prime}})= \prod_{i=1}^{N}\langle x_{i};z,\alpha | 
 x_{i}^{\prime};z,\alpha \rangle.
 \end{eqnarray}
In the case of $\alpha-SU(2)$ feature space, the kernel is obtained as follows:\\
\begin{tcolorbox}[boxsep=2pt,left=2pt,right=2pt,top=4pt,bottom=4pt]
\bf{$\bf{\alpha-SU(2)}$ Kernel Function}
\begin{eqnarray}
K(\mathbf{x},\mathbf{x^{\prime}})= \prod_{i=1}^{N} \left[\frac{1+\tan^{2}\left(z\sqrt{\frac{|\alpha|}{2}}\right)e^{i(x_{i}-x^{\prime}_{i})}}{1+\tan^{2}\left(z\sqrt{\frac{|\alpha|}{2}}\right)}\right]^{2k},
\end{eqnarray}
\end{tcolorbox}
Moreover, in the case of $\alpha-SU(1,1)$ feature space, the kernel is given as follows:
\begin{tcolorbox}[boxsep=2pt,left=2pt,right=2pt,top=4pt,bottom=4pt]
\bf{$\bf{\alpha-SU(1,1)}$ Kernel Function}
\begin{eqnarray}
K(\mathbf{x},\mathbf{x^{\prime}})= \prod_{i=1}^{N} \left[\frac{1-\tanh^{2}\left(z\sqrt{\frac{|\alpha|}{2}}\right)}{1-\tanh^{2}\left(z\sqrt{\frac{|\alpha|}{2}}\right)e^{i(x_{i}-x^{\prime}_{i})}}\right]^{2k}.
\end{eqnarray}
\end{tcolorbox}
\indent For understanding the role of $\alpha$ and $k$, we study the geometrical properties of the above-mentioned feature spaces.  We can define the line element of the feature space, by using the Fubini–Study metric \cite{bengtsson2017geometry}, that is,
\begin{eqnarray}
ds^{2}=\| d|x;z,\alpha\rangle\|^{2}-|\langle x;z,\alpha|d|x;z,\alpha\rangle|^{2},
\end{eqnarray}
 By using the above definition, the metric of $\alpha-SU(2)$ feature space is obtained by
\begin{eqnarray}\label{eq11}
ds^{2}=k\alpha dz^{2}+\frac{k}{2} \sin^{2} \left(z\sqrt{2\alpha}\right)dx^{2},
\end{eqnarray}
which describes a positive constant curvature with the scalar Ricci $R=4/k$. This is in fact a surface of revolution conforming with a sphere \cite{carinena2005central}.
By using the same method, the metric of the $\alpha-SU(1,1)$ feature space is given by
\begin{eqnarray}\label{eq10}
ds^{2}=k|\alpha| dz^{2}+\frac{k}{2} \sinh^{2} \left(z\sqrt{2|\alpha|}\right)dx^{2}.
\end{eqnarray}
which describes a negative constant curvature with the scalar Ricci $R=-4/k$, conformal with pseudo-spheres \cite{carinena2005central}. Figure \ref{parametric_feature_spaces} shows topological categories of feature spaces associated with $\alpha-SU(2)$ and $\alpha-SU(1,1)$ coherent states.
As seen above, two differing categories of the deformed coherent states are defined with different topologies: one ($\alpha-SU(2)$) is constructed on a truncated Hilbert space, forming a compact constant curvature feature space, conforming with a sphere. The other ($\alpha-SU(1,1)$) is built on an infinite Hilbert space that leads to a feature space conforming with a pseudo-sphere, with negative constant curvature.

In the next section, we present a illustrate the empirical effectiveness of the meta-theoretically derived kernel functions in different experimental settings.

\section{Empirical evaluation of the kernel functions}\label{test_design}

In order to assess the effectiveness of the proposed kernel functions, we conducted a series of experiments using different well known synthetic datasets of the literature, namely 1) Python's \textit{scikit-learn} library: the circles, moons, and 2) the iris dataset. 
The circles and the moons dataset involves binary features in a binary classification problem. The iris dataset is a multiclass classification task with three classes and four features.

\begin{figure}[t]
    \centering
    \includegraphics[scale=.6]{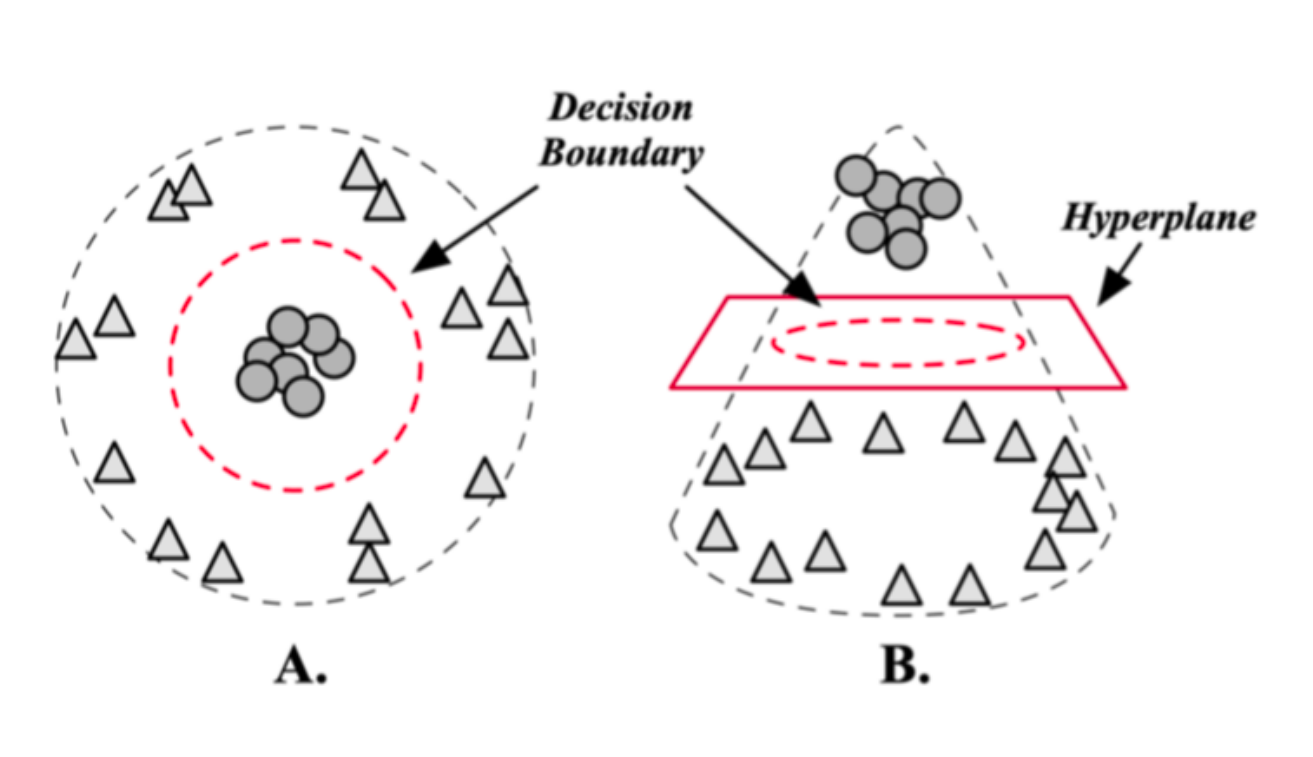}
    \caption{Kernel Trick. A kernel is applied to each data point to map the original non-linear observations into a higher dimensional space where the observations may become linearly separable through a hyperplane.}
    \label{fig:svm_hyper}
\end{figure}

\begin{figure}[t]
    \centering
  \includegraphics[scale=0.55]{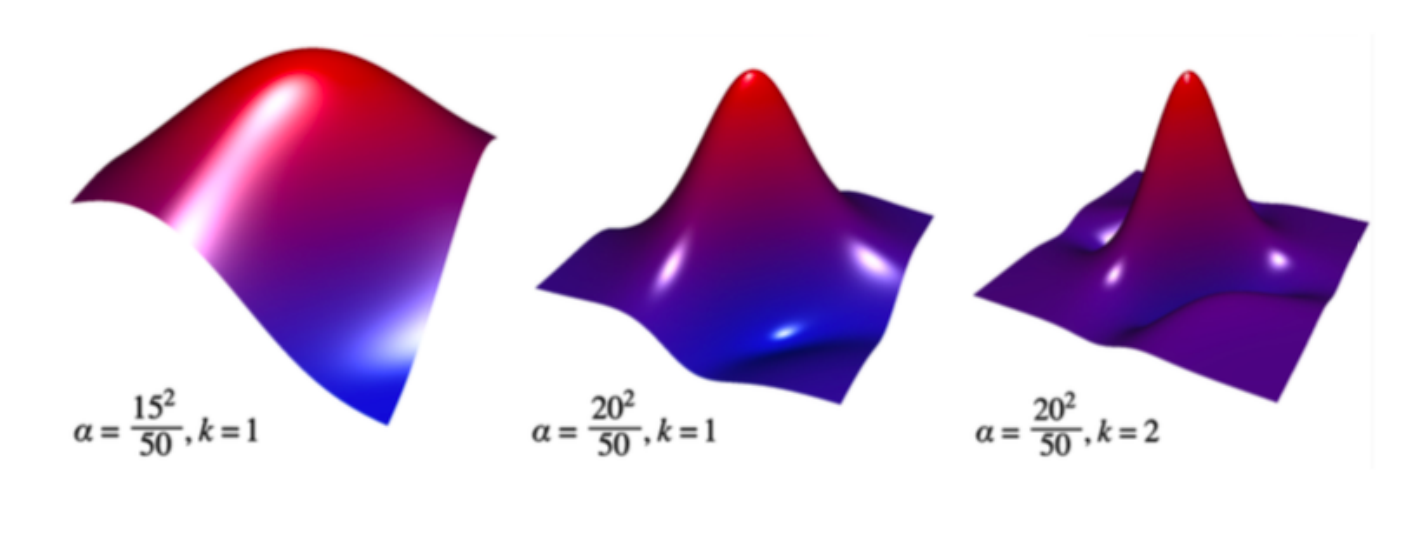}
  \caption{Shape of the  kernel function obtained by $\alpha-SU(1,1)$-coherent state  for different strength hyperparameters $\alpha$ and $k$, while $z=1$. The input $x$ is fixed at $(0, 0)$ and $x^{\prime}$ is varied.}
  \label{vis_su11}
\end{figure}

\begin{figure}[t]
\centering
  \includegraphics[scale=.55]{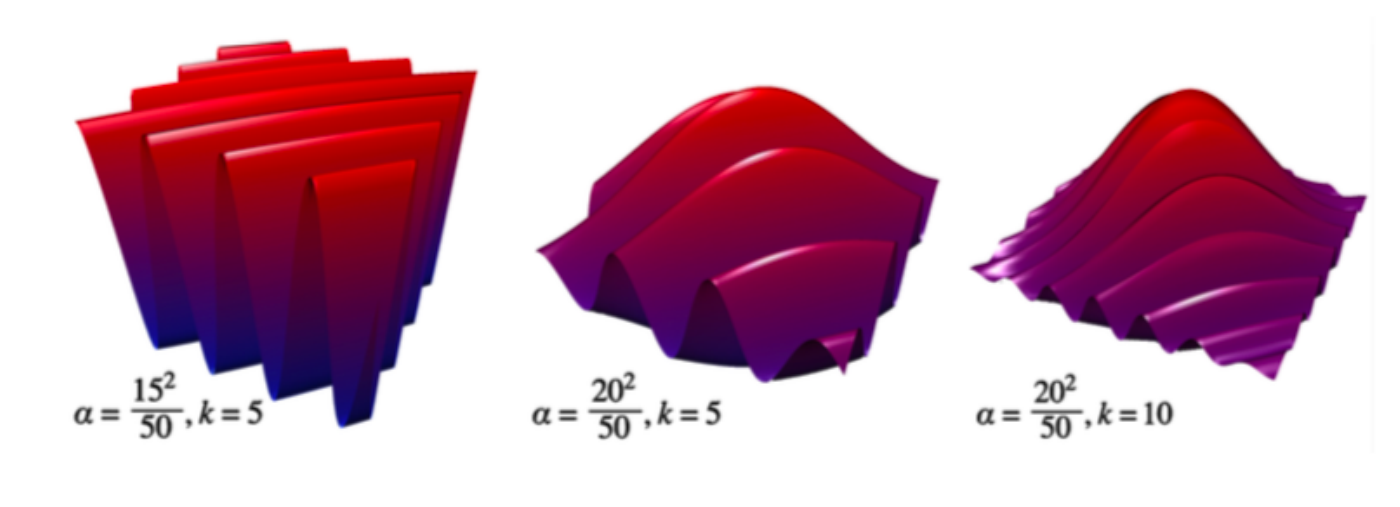}
  \caption{Shape of the  kernel function obtained by $\alpha-SU(2)$-coherent state  for different strength hyperparameters $\alpha$ and $k$, while $z=1$. The input $x$ is fixed at $(0, 0)$ and $x^{\prime}$ is varied.}
  \label{fig:vis_su2}
\end{figure}

Since the goal of SVMs is to find the maximum-margin hyperplane, a set of parameters are needed to control the error between these margins (Figure~\ref{fig:svm_hyper} shows this optimization in the high-dimensional feature space). For the RBF kernel, two parameters play a major role in this optimization process:
\begin{itemize}
\item Hyperparameter $C$: is a regularization parameter that controls the trade-off between the decision boundary and mis-classification term. It basically controls how much mis-classifications are tolerable during the optimization problem.
\item $\gamma$, which controls the non-linearity of the decision boundary. It defines how far influences the calculation of plausible line of separation. A low $\gamma$ takes into consideration far away points to influence the decision boundary; a high $\gamma$ considers only points that are close to the decision boundary.
\end{itemize}
When considering the proposed meta-kernel, we have a set of new parameters that extend the current RBF kernel with a set of new non-linear functions that are based on the  dW-H algebra. These extra parameters also enable the construction of a feature space over a surface of revolution with constant curvature. Theoretically, this could lead to significant improvements when the dataset is distributed along revolution surfaces. The parameters for both $SU(1,1)$ and $SU(2)$ kernels are the following:
\begin{itemize}
    \item Parameter $k$ is related to the curvature of the feature surface and  controls the non-linearity of the decision boundary in such way that high values of the parameter consider points that are near to the decision boundary whilst low values cause points further away to influence the decision boundary. Figures \ref{vis_su11} and \ref{fig:vis_su2} illustrate the rule of the parameter $k$.   
    \item By considering a fixed curvature, i.e., $k=const.$, the product of parameters $z$ and $\sqrt{\alpha}$, as an extra parameter   $z\sqrt{\alpha}$, controls the decision boundary as well. Figure   \ref{vis_su11} and \ref{fig:vis_su2} indicates a schematic behaviour of parameters $\alpha$ and $k$, for $\alpha-SU(1,1)$ and $\alpha-SU(2)$ respectively.  
\end{itemize}

\begin{figure}[t]
\centering
  \includegraphics[scale=.7]{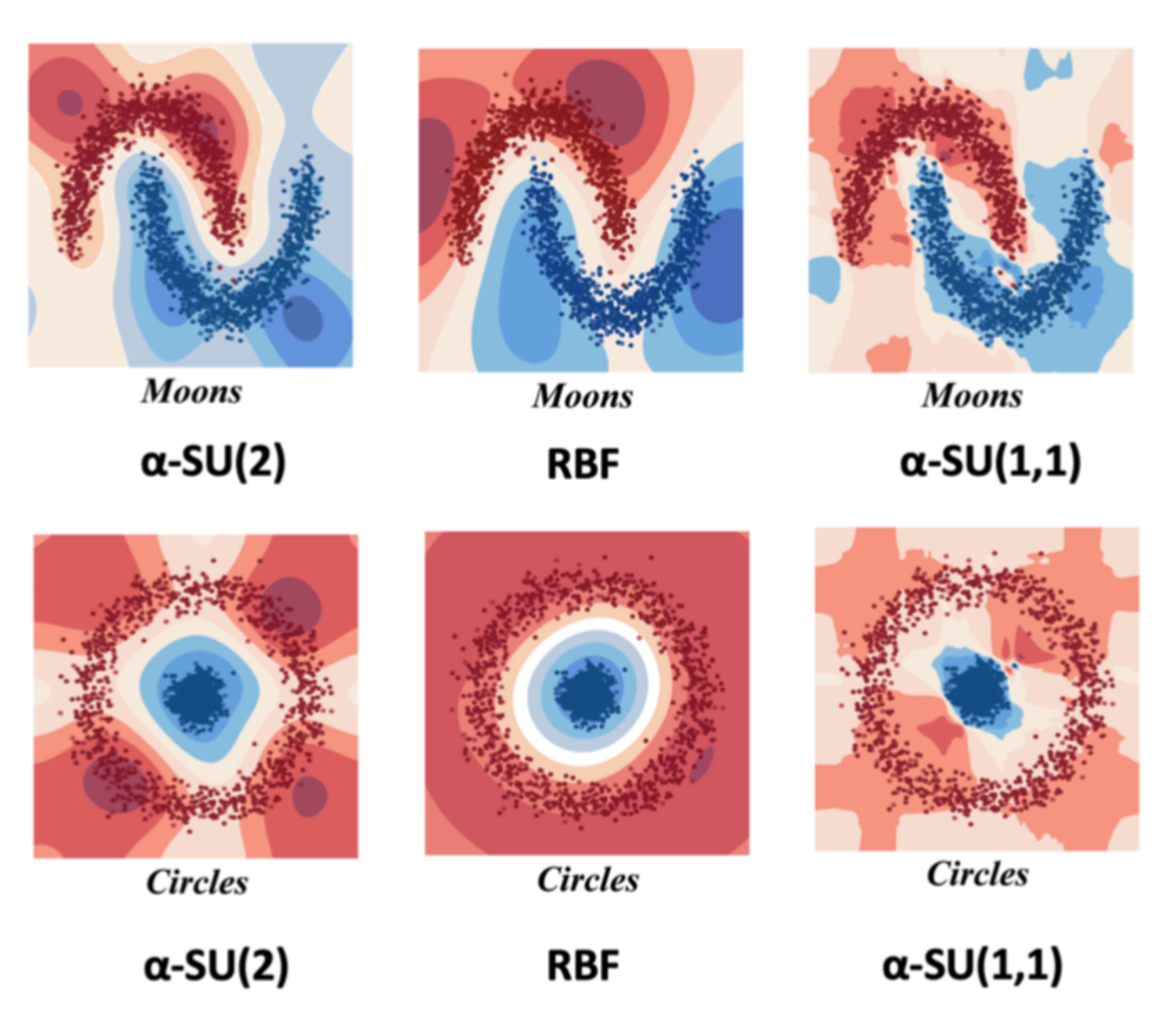}
  \caption{Decision boundaries computed using different kernels for the synthetic datasets \textit{Moons} and \textit{Circles}.}
  \label{fig:svc_classification}
\end{figure}

Figure~\ref{fig:svc_classification} illustrates different decision boundaries that can be computed using the different kernels. One can see the different non-linearity properties of the $\alpha-SU(2)$ and $\alpha-SU(1,1)$ kernels compared to the standard RBF kernel.

\begin{figure}[]
\centering
  \includegraphics[scale=.7]{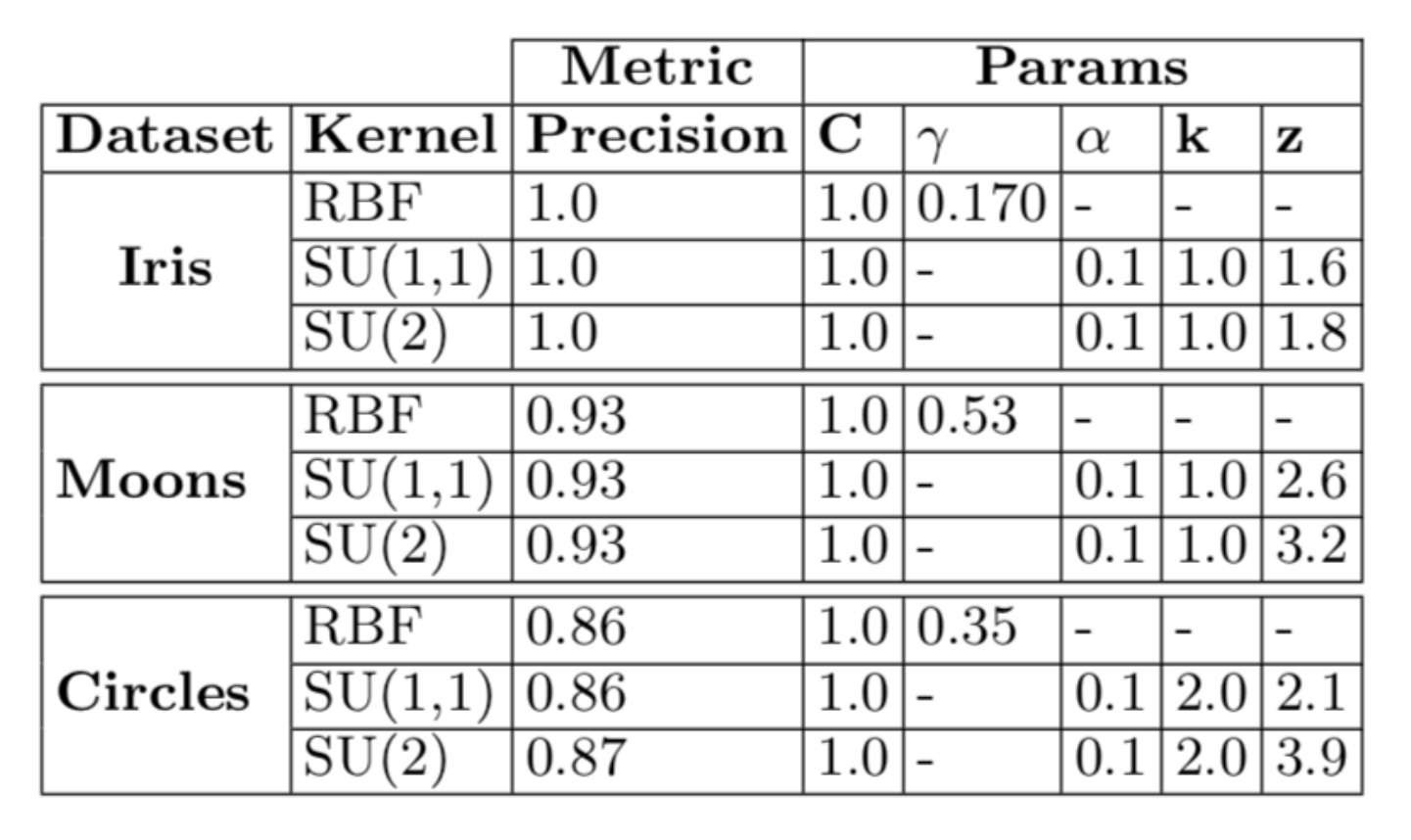}
  \caption{Results obtained using the proposed kernels $\alpha$-$SU(1,1)$ and $\alpha$-$SU(2)$ for different datasets. Note that for the Moons and Circles synthetic dataset, we generated 1000 samples of which $70\%$ were used for training and the remaining $30\%$ for test. These datasets were generated using a noise factor of 0.3 and 0.1, respectively.}
\label{tab:results}
\end{figure}

For evaluation purposes, the parameters that provided the best results in terms of precision were found using a grid search approach (for more details on the evaluation code and experimental setup. For the synthetic datasets, Moons and Circles, 1000 samples were generated, where 70\% were used for the training process and 30\% were used for the evaluation task. To make the classification task more challenging, we applied noise factors of 0.3 and 0.1 to these datasets, respectively. Learning curves were analysed to ensure unbiased results and no overfitting.  Table~\ref{tab:results} summarises the results.

 The proposed meta-kernels $\alpha$-SU(1,1) and $\alpha$-SU(2) exhibit state of the art performance when compared with the RBF kernel. 
 These kernels provide a significant advantage for data points distributed over curved surfaces. Given that it is hard to find benchmark datasets with those characteristics, the results presented in Table~\ref{tab:results} suggest an cautiously encouraging first step.


\section{Discussion} \label{discussion}

In SVM-based classification problems, the appropriate choice of a kernel is fundamental to achieve high classification performance. 
However, the current 'trial-and-error' nature of selecting the best kernel poses significant challenges, especially when one considers kernels that can support both classical and quantum-inspired machine learning algorithms~\cite{Ali06}. 
In this section, a visual analysis is provided of how different kernels are derived from the $\alpha$ and $k$ parameters of both $\alpha-SU(1,1)$ and $\alpha-SU(2)$ kernels in order to promote a clear discrimination between kernel functions. 

The $\alpha$ parameter controls allows a specific kernel function is derived from  deformed Weyl-Heisenberg (dW-H) algebra. 
A value of $\alpha=0$ is a flat surface. When $\alpha$ is small, it contributes to an almost flat surface, when maintaining $k$ low. On the other hand, when $\alpha$ is high, it contributes to squeeze the function towards its center. Figure~\ref{vis_su11} shows the impact of parameter $\alpha$ in in the $\alpha-SU(1,1)$ kernel.
Regarding the $\alpha-SU(2)$ kernels, the parameter $\alpha$ generates a kernel that maps the non-linear observations into a higher dimensional space that `folds' the data in the feature space. A high value in $\alpha$ and $k$ squeeze  these folds towards the center of the distribution of the geodesic distances between the data points as visualized in Figure~\ref{fig:vis_su2}.

In terms of the empirical evaluation of the kernels, Table~\ref{tab:results} indicates the the best results were obtained with low values of $\alpha = 0.1$, $1.0 \leq k \leq 2.0$, with the $z$ parameters in the range: $ 0.2 \leq z \leq 4.6$. 
\section{Conclusion}
In this paper, 
by using the theory of non-linear coherent states, we put forward a meta-kernel approach for deriving kernel functions for use in ML. 
More specifically, data is mapped into a feature space which is defined as a deformed coherent state as defined by a deformed Weyl-Heisenberg algebra.
This algebra unifies the well-known $SU(2)$, Weyl-Heisenberg, and $SU(1,1)$ groups, through a common  parameter $\alpha$.
In addition, by studying tgeometrical properties of feature space constructed on the dW-H coherent state, we showed that the meta-kernel function applies associated  surfaces of revolution as feature spaces identified with non-linear coherent states.  
An empirical investigation compares the $\alpha-SU(2)$ and $\alpha-SU(1,1)$ kernels derived from the meta-kernel which shows performance similar to the Radial Basis kernel.

Kernel functions drive developments in the field of machine learning and the meta-kernel function presented in this paper opens new theoretical avenues for the definition and exploration of kernel functions.\\

\section*{Acknowledgement}
This research was supported by the Asian Office of Aerospace Research and Development (AOARD) grant: FA2386-17-1-4016.

\appendices
\section{Geometrical Properties of Feature surfaces}
The  Christoffel symbols of the second kind according to definition are give by
\begin{eqnarray}
\Gamma_{ij}^{k}=\frac{1}{2}g^{kl}\left[\partial_{i}g_{jl}+\partial_{j}g_{il}-\partial_{l}g_{ij}\right]
\end{eqnarray}
in which $g_{ij}$ is the $(i,j)$th component of the metric, $g^{ij}=g_{ij}^{-1}$ and $\partial_{i}$ is an abbreviation of $\frac{\partial}{\partial x_{i}}$. Also, according to standard notation, the Einstein summation convention is applied, i.e.,  summation over a set of indexed terms in a formula, e.g. $g^{ij}g_{jk}=g^{i1}g_{1k}+g^{i2}g_{2k}$. By using the metric (\ref{eq11}), non-zero components of 
the  Christoffel symbols of the second kind are respectively given by:
\begin{eqnarray}
\Gamma_{xz}^{x}&=&\sqrt{2\alpha}\cot (\sqrt{2\alpha}\ z)\nonumber\\
\Gamma_{xx}^{z}&=&-\frac{1}{2\sqrt{2\alpha}}\sin (2\sqrt{2\alpha}\ z)
\end{eqnarray}
Also, according to definition, the Ricci tensor is given by
\begin{eqnarray}
R_{ij}=\partial_{k}\Gamma_{ij}^{k}-\partial_{i}\Gamma_{kj}^{k}+
\Gamma_{ij}^{k}\Gamma_{kl}^{l}-
\Gamma_{ik}^{l}\Gamma_{lj}^{k}.
\end{eqnarray}{}
Hence,
the non-zero Ricci tensors are given by:
\begin{eqnarray}
R_{xx}= \sin^{2} \left[\sqrt{2\alpha} z\right], R_{zz}=2\alpha. 
\end{eqnarray}
The Ricci scalar, which gives the curvature, is obtained by
\begin{eqnarray}
R=\frac{4}{k}
\end{eqnarray}
\indent In the case of metric (\ref{eq10}), non-zero components are given by
\begin{eqnarray}
\Gamma_{xz}^{x}&=&\sqrt{2\alpha}\coth (\sqrt{2\alpha}\ z)\nonumber\\
\Gamma_{xx}^{z}&=&\frac{-1}{2\sqrt{2\alpha}}\sinh (2\sqrt{2\alpha}\ z)
\end{eqnarray}
and the Ricci tensor are given by
\begin{eqnarray}
R_{xx}=\sinh^{2}(z\sqrt{2\alpha}), R_{zz}=-2\alpha 
\end{eqnarray}
The Ricci scalar, which gives the curvature, is obtained by
\begin{eqnarray}
R=-\frac{4}{k}
\end{eqnarray}

\bibliographystyle{unsrt} 
\bibliography{apssamp}

\begin{thebibliography}{37}%
\makeatletter
\providecommand \@ifxundefined [1]{%
 \@ifx{#1\undefined}
}%
\providecommand \@ifnum [1]{%
 \ifnum #1\expandafter \@firstoftwo
 \else \expandafter \@secondoftwo
 \fi
}%
\providecommand \@ifx [1]{%
 \ifx #1\expandafter \@firstoftwo
 \else \expandafter \@secondoftwo
 \fi
}%
\providecommand \natexlab [1]{#1}%
\providecommand \enquote  [1]{``#1''}%
\providecommand \bibnamefont  [1]{#1}%
\providecommand \bibfnamefont [1]{#1}%
\providecommand \citenamefont [1]{#1}%
\providecommand \href@noop [0]{\@secondoftwo}%
\providecommand \href [0]{\begingroup \@sanitize@url \@href}%
\providecommand \@href[1]{\@@startlink{#1}\@@href}%
\providecommand \@@href[1]{\endgroup#1\@@endlink}%
\providecommand \@sanitize@url [0]{\catcode `\\12\catcode `\$12\catcode
  `\&12\catcode `\#12\catcode `\^12\catcode `\_12\catcode `\%12\relax}%
\providecommand \@@startlink[1]{}%
\providecommand \@@endlink[0]{}%
\providecommand \url  [0]{\begingroup\@sanitize@url \@url }%
\providecommand \@url [1]{\endgroup\@href {#1}{\urlprefix }}%
\providecommand \urlprefix  [0]{URL }%
\providecommand \Eprint [0]{\href }%
\providecommand \doibase [0]{https://doi.org/}%
\providecommand \selectlanguage [0]{\@gobble}%
\providecommand \bibinfo  [0]{\@secondoftwo}%
\providecommand \bibfield  [0]{\@secondoftwo}%
\providecommand \translation [1]{[#1]}%
\providecommand \BibitemOpen [0]{}%
\providecommand \bibitemStop [0]{}%
\providecommand \bibitemNoStop [0]{.\EOS\space}%
\providecommand \EOS [0]{\spacefactor3000\relax}%
\providecommand \BibitemShut  [1]{\csname bibitem#1\endcsname}%
\let\auto@bib@innerbib\@empty
\bibitem [{\citenamefont {Boser}\ \emph {et~al.}(1992)\citenamefont {Boser},
  \citenamefont {Guyon},\ and\ \citenamefont {Vapnik}}]{boser1992training}%
  \BibitemOpen
  \bibfield  {author} {\bibinfo {author} {\bibfnamefont {B.~E.}\ \bibnamefont
  {Boser}}, \bibinfo {author} {\bibfnamefont {I.~M.}\ \bibnamefont {Guyon}},\
  and\ \bibinfo {author} {\bibfnamefont {V.~N.}\ \bibnamefont {Vapnik}},\
  }\bibfield  {title} {\bibinfo {title} {A training algorithm for optimal
  margin classifiers},\ }in\ \href@noop {} {\emph {\bibinfo {booktitle}
  {Proceedings of the fifth annual workshop on Computational learning
  theory}}}\ (\bibinfo {organization} {ACM},\ \bibinfo {year} {1992})\ pp.\
  \bibinfo {pages} {144--152}\BibitemShut {NoStop}%
\bibitem [{\citenamefont {Cortes}\ and\ \citenamefont
  {Vapnik}(1995)}]{cortes1995support}%
  \BibitemOpen
  \bibfield  {author} {\bibinfo {author} {\bibfnamefont {C.}~\bibnamefont
  {Cortes}}\ and\ \bibinfo {author} {\bibfnamefont {V.}~\bibnamefont
  {Vapnik}},\ }\bibfield  {title} {\bibinfo {title} {Support-vector networks},\
  }\href@noop {} {\bibfield  {journal} {\bibinfo  {journal} {Machine learning}\
  }\textbf {\bibinfo {volume} {20}},\ \bibinfo {pages} {273} (\bibinfo {year}
  {1995})}\BibitemShut {NoStop}%
\bibitem [{\citenamefont {Haykin}(2008)}]{Haykin08}%
  \BibitemOpen
  \bibfield  {author} {\bibinfo {author} {\bibfnamefont {S.~O.}\ \bibnamefont
  {Haykin}},\ }\href@noop {} {\emph {\bibinfo {title} {Neural Networks and
  Learning Machines: A Comprehensive Foundation}}}\ (\bibinfo  {publisher}
  {Pearson (3rd edition)},\ \bibinfo {year} {2008})\BibitemShut {NoStop}%
\bibitem [{\citenamefont {Saugat~Bhattacharyya}(2011)}]{Bhattacharyya11}%
  \BibitemOpen
  \bibfield  {author} {\bibinfo {author} {\bibfnamefont {A.~K. D. T. A. K.~N.}\
  \bibnamefont {Saugat~Bhattacharyya}, \bibfnamefont {Anwesha~Khasnobish}},\
  }\bibfield  {title} {\bibinfo {title} {Performance analysis of left/right
  hand movement classification from eeg signal by intelligent algorithms},\
  }in\ \href@noop {} {\emph {\bibinfo {booktitle} {IEEE Symposium on
  Computational Intelligence, Cognitive Algorithms, Mind, and Brain (CCMB)}}}\
  (\bibinfo {year} {2011})\ pp.\ \bibinfo {pages} {1--8}\BibitemShut {NoStop}%
\bibitem [{\citenamefont {Schuld}\ and\ \citenamefont
  {Killoran}(2019)}]{schuld2019quantum}%
  \BibitemOpen
  \bibfield  {author} {\bibinfo {author} {\bibfnamefont {M.}~\bibnamefont
  {Schuld}}\ and\ \bibinfo {author} {\bibfnamefont {N.}~\bibnamefont
  {Killoran}},\ }\bibfield  {title} {\bibinfo {title} {Quantum machine learning
  in feature hilbert spaces},\ }\href@noop {} {\bibfield  {journal} {\bibinfo
  {journal} {Physical review letters}\ }\textbf {\bibinfo {volume} {122}},\
  \bibinfo {pages} {040504} (\bibinfo {year} {2019})}\BibitemShut {NoStop}%
\bibitem [{\citenamefont {Schuld}(2019)}]{schuld2019machine}%
  \BibitemOpen
  \bibfield  {author} {\bibinfo {author} {\bibfnamefont {M.}~\bibnamefont
  {Schuld}},\ }\href@noop {} {\bibinfo {title} {Machine learning in quantum
  spaces}} (\bibinfo {year} {2019})\BibitemShut {NoStop}%
\bibitem [{\citenamefont {Havl{\'\i}{\v{c}}ek}\ \emph
  {et~al.}(2019)\citenamefont {Havl{\'\i}{\v{c}}ek}, \citenamefont
  {C{\'o}rcoles}, \citenamefont {Temme}, \citenamefont {Harrow}, \citenamefont
  {Kandala}, \citenamefont {Chow},\ and\ \citenamefont
  {Gambetta}}]{havlivcek2019supervised}%
  \BibitemOpen
  \bibfield  {author} {\bibinfo {author} {\bibfnamefont {V.}~\bibnamefont
  {Havl{\'\i}{\v{c}}ek}}, \bibinfo {author} {\bibfnamefont {A.~D.}\
  \bibnamefont {C{\'o}rcoles}}, \bibinfo {author} {\bibfnamefont
  {K.}~\bibnamefont {Temme}}, \bibinfo {author} {\bibfnamefont {A.~W.}\
  \bibnamefont {Harrow}}, \bibinfo {author} {\bibfnamefont {A.}~\bibnamefont
  {Kandala}}, \bibinfo {author} {\bibfnamefont {J.~M.}\ \bibnamefont {Chow}},\
  and\ \bibinfo {author} {\bibfnamefont {J.~M.}\ \bibnamefont {Gambetta}},\
  }\bibfield  {title} {\bibinfo {title} {Supervised learning with
  quantum-enhanced feature spaces},\ }\href@noop {} {\bibfield  {journal}
  {\bibinfo  {journal} {Nature}\ }\textbf {\bibinfo {volume} {567}},\ \bibinfo
  {pages} {209} (\bibinfo {year} {2019})}\BibitemShut {NoStop}%
\bibitem [{\citenamefont {Mehta}\ \emph {et~al.}(2019)\citenamefont {Mehta},
  \citenamefont {Bukov}, \citenamefont {Wang}, \citenamefont {Day},
  \citenamefont {Richardson}, \citenamefont {Fisher},\ and\ \citenamefont
  {Schwab}}]{mehta2019high}%
  \BibitemOpen
  \bibfield  {author} {\bibinfo {author} {\bibfnamefont {P.}~\bibnamefont
  {Mehta}}, \bibinfo {author} {\bibfnamefont {M.}~\bibnamefont {Bukov}},
  \bibinfo {author} {\bibfnamefont {C.-H.}\ \bibnamefont {Wang}}, \bibinfo
  {author} {\bibfnamefont {A.~G.}\ \bibnamefont {Day}}, \bibinfo {author}
  {\bibfnamefont {C.}~\bibnamefont {Richardson}}, \bibinfo {author}
  {\bibfnamefont {C.~K.}\ \bibnamefont {Fisher}},\ and\ \bibinfo {author}
  {\bibfnamefont {D.~J.}\ \bibnamefont {Schwab}},\ }\bibfield  {title}
  {\bibinfo {title} {A high-bias, low-variance introduction to machine learning
  for physicists},\ }\href@noop {} {\bibfield  {journal} {\bibinfo  {journal}
  {Physics reports}\ } (\bibinfo {year} {2019})}\BibitemShut {NoStop}%
\bibitem [{\citenamefont {Killoran}\ \emph {et~al.}(2019)\citenamefont
  {Killoran}, \citenamefont {Izaac}, \citenamefont {Quesada}, \citenamefont
  {Bergholm}, \citenamefont {Amy},\ and\ \citenamefont
  {Weedbrook}}]{killoran2019strawberry}%
  \BibitemOpen
  \bibfield  {author} {\bibinfo {author} {\bibfnamefont {N.}~\bibnamefont
  {Killoran}}, \bibinfo {author} {\bibfnamefont {J.}~\bibnamefont {Izaac}},
  \bibinfo {author} {\bibfnamefont {N.}~\bibnamefont {Quesada}}, \bibinfo
  {author} {\bibfnamefont {V.}~\bibnamefont {Bergholm}}, \bibinfo {author}
  {\bibfnamefont {M.}~\bibnamefont {Amy}},\ and\ \bibinfo {author}
  {\bibfnamefont {C.}~\bibnamefont {Weedbrook}},\ }\bibfield  {title} {\bibinfo
  {title} {Strawberry fields: A software platform for photonic quantum
  computing},\ }\href@noop {} {\bibfield  {journal} {\bibinfo  {journal}
  {Quantum}\ }\textbf {\bibinfo {volume} {3}},\ \bibinfo {pages} {129}
  (\bibinfo {year} {2019})}\BibitemShut {NoStop}%
\bibitem [{\citenamefont {Killoran}\ \emph {et~al.}(2018)\citenamefont
  {Killoran}, \citenamefont {Bromley}, \citenamefont {Arrazola}, \citenamefont
  {Schuld}, \citenamefont {Quesada},\ and\ \citenamefont
  {Lloyd}}]{killoran2018continuous}%
  \BibitemOpen
  \bibfield  {author} {\bibinfo {author} {\bibfnamefont {N.}~\bibnamefont
  {Killoran}}, \bibinfo {author} {\bibfnamefont {T.~R.}\ \bibnamefont
  {Bromley}}, \bibinfo {author} {\bibfnamefont {J.~M.}\ \bibnamefont
  {Arrazola}}, \bibinfo {author} {\bibfnamefont {M.}~\bibnamefont {Schuld}},
  \bibinfo {author} {\bibfnamefont {N.}~\bibnamefont {Quesada}},\ and\ \bibinfo
  {author} {\bibfnamefont {S.}~\bibnamefont {Lloyd}},\ }\bibfield  {title}
  {\bibinfo {title} {Continuous-variable quantum neural networks},\ }\href@noop
  {} {\bibfield  {journal} {\bibinfo  {journal} {arXiv preprint
  arXiv:1806.06871}\ } (\bibinfo {year} {2018})}\BibitemShut {NoStop}%
\bibitem [{\citenamefont {Srinivasan}\ \emph {et~al.}(2018)\citenamefont
  {Srinivasan}, \citenamefont {Downey},\ and\ \citenamefont
  {Boots}}]{srinivasan2018learning}%
  \BibitemOpen
  \bibfield  {author} {\bibinfo {author} {\bibfnamefont {S.}~\bibnamefont
  {Srinivasan}}, \bibinfo {author} {\bibfnamefont {C.}~\bibnamefont {Downey}},\
  and\ \bibinfo {author} {\bibfnamefont {B.}~\bibnamefont {Boots}},\ }\bibfield
   {title} {\bibinfo {title} {Learning and inference in hilbert space with
  quantum graphical models},\ }in\ \href@noop {} {\emph {\bibinfo {booktitle}
  {Advances in Neural Information Processing Systems}}}\ (\bibinfo {year}
  {2018})\ pp.\ \bibinfo {pages} {10338--10347}\BibitemShut {NoStop}%
\bibitem [{\citenamefont {Adhikary}\ \emph {et~al.}(2019)\citenamefont
  {Adhikary}, \citenamefont {Dangwal},\ and\ \citenamefont
  {Bhowmik}}]{adhikary2019supervised}%
  \BibitemOpen
  \bibfield  {author} {\bibinfo {author} {\bibfnamefont {S.}~\bibnamefont
  {Adhikary}}, \bibinfo {author} {\bibfnamefont {S.}~\bibnamefont {Dangwal}},\
  and\ \bibinfo {author} {\bibfnamefont {D.}~\bibnamefont {Bhowmik}},\
  }\bibfield  {title} {\bibinfo {title} {Supervised learning with a quantum
  classifier using a multi-level system},\ }\href@noop {} {\bibfield  {journal}
  {\bibinfo  {journal} {arXiv preprint arXiv:1908.08385}\ } (\bibinfo {year}
  {2019})}\BibitemShut {NoStop}%
\bibitem [{\citenamefont {Bartkiewicz}\ \emph {et~al.}(2019)\citenamefont
  {Bartkiewicz}, \citenamefont {Gneiting}, \citenamefont {{\v{C}}ernoch},
  \citenamefont {Jir{\'a}kov{\'a}}, \citenamefont {Lemr},\ and\ \citenamefont
  {Nori}}]{bartkiewicz2019experimental}%
  \BibitemOpen
  \bibfield  {author} {\bibinfo {author} {\bibfnamefont {K.}~\bibnamefont
  {Bartkiewicz}}, \bibinfo {author} {\bibfnamefont {C.}~\bibnamefont
  {Gneiting}}, \bibinfo {author} {\bibfnamefont {A.}~\bibnamefont
  {{\v{C}}ernoch}}, \bibinfo {author} {\bibfnamefont {K.}~\bibnamefont
  {Jir{\'a}kov{\'a}}}, \bibinfo {author} {\bibfnamefont {K.}~\bibnamefont
  {Lemr}},\ and\ \bibinfo {author} {\bibfnamefont {F.}~\bibnamefont {Nori}},\
  }\bibfield  {title} {\bibinfo {title} {Experimental kernel-based quantum
  machine learning in finite feature space},\ }\href@noop {} {\bibfield
  {journal} {\bibinfo  {journal} {arXiv preprint arXiv:1906.04137}\ } (\bibinfo
  {year} {2019})}\BibitemShut {NoStop}%
\bibitem [{\citenamefont {K{\"u}bler}\ \emph {et~al.}(2019)\citenamefont
  {K{\"u}bler}, \citenamefont {Muandet},\ and\ \citenamefont
  {Sch{\"o}lkopf}}]{kubler2019quantum}%
  \BibitemOpen
  \bibfield  {author} {\bibinfo {author} {\bibfnamefont {J.~M.}\ \bibnamefont
  {K{\"u}bler}}, \bibinfo {author} {\bibfnamefont {K.}~\bibnamefont
  {Muandet}},\ and\ \bibinfo {author} {\bibfnamefont {B.}~\bibnamefont
  {Sch{\"o}lkopf}},\ }\bibfield  {title} {\bibinfo {title} {Quantum mean
  embedding of probability distributions},\ }\href@noop {} {\bibfield
  {journal} {\bibinfo  {journal} {Physical Review Research}\ }\textbf {\bibinfo
  {volume} {1}},\ \bibinfo {pages} {033159} (\bibinfo {year}
  {2019})}\BibitemShut {NoStop}%
\bibitem [{\citenamefont {Combescure}\ and\ \citenamefont
  {Robert}(2012)}]{combescure2012coherent}%
  \BibitemOpen
  \bibfield  {author} {\bibinfo {author} {\bibfnamefont {M.}~\bibnamefont
  {Combescure}}\ and\ \bibinfo {author} {\bibfnamefont {D.}~\bibnamefont
  {Robert}},\ }\href@noop {} {\emph {\bibinfo {title} {Coherent states and
  applications in mathematical physics}}}\ (\bibinfo  {publisher} {Springer
  Science \& Business Media},\ \bibinfo {year} {2012})\BibitemShut {NoStop}%
\bibitem [{\citenamefont {Ali}\ \emph {et~al.}(2000)\citenamefont {Ali},
  \citenamefont {Antoine}, \citenamefont {Gazeau} \emph
  {et~al.}}]{ali2000coherent}%
  \BibitemOpen
  \bibfield  {author} {\bibinfo {author} {\bibfnamefont {S.~T.}\ \bibnamefont
  {Ali}}, \bibinfo {author} {\bibfnamefont {J.-P.}\ \bibnamefont {Antoine}},
  \bibinfo {author} {\bibfnamefont {J.-P.}\ \bibnamefont {Gazeau}}, \emph
  {et~al.},\ }\href@noop {} {\emph {\bibinfo {title} {Coherent states, wavelets
  and their generalizations}}},\ Vol.~\bibinfo {volume} {3}\ (\bibinfo
  {publisher} {Springer},\ \bibinfo {year} {2000})\BibitemShut {NoStop}%
\bibitem [{\citenamefont {Ali}\ and\ \citenamefont
  {A.Smith-Miles}(2006)}]{Ali06}%
  \BibitemOpen
  \bibfield  {author} {\bibinfo {author} {\bibfnamefont {S.}~\bibnamefont
  {Ali}}\ and\ \bibinfo {author} {\bibfnamefont {K.}~\bibnamefont
  {A.Smith-Miles}},\ }\bibfield  {title} {\bibinfo {title} {A meta-learning
  approach to automatic kernel selection for support vector machines},\
  }\href@noop {} {\bibfield  {journal} {\bibinfo  {journal} {Neurocomputing}\
  }\textbf {\bibinfo {volume} {70}},\ \bibinfo {pages} {173} (\bibinfo {year}
  {2006})}\BibitemShut {NoStop}%
\bibitem [{\citenamefont {Dehdashti}\ \emph {et~al.}(2013)\citenamefont
  {Dehdashti}, \citenamefont {Mahdifar},\ and\ \citenamefont
  {Roknizadeh}}]{dehdashti2013coherent}%
  \BibitemOpen
  \bibfield  {author} {\bibinfo {author} {\bibfnamefont {S.}~\bibnamefont
  {Dehdashti}}, \bibinfo {author} {\bibfnamefont {A.}~\bibnamefont
  {Mahdifar}},\ and\ \bibinfo {author} {\bibfnamefont {R.}~\bibnamefont
  {Roknizadeh}},\ }\bibfield  {title} {\bibinfo {title} {Coherent state of
  $\alpha$-deformed weyl--heisenberg algebra},\ }\href@noop {} {\bibfield
  {journal} {\bibinfo  {journal} {International Journal of Geometric Methods in
  Modern Physics}\ }\textbf {\bibinfo {volume} {10}},\ \bibinfo {pages}
  {1350014} (\bibinfo {year} {2013})}\BibitemShut {NoStop}%
\bibitem [{\citenamefont {Dehdashti}\ \emph {et~al.}(2015)\citenamefont
  {Dehdashti}, \citenamefont {Li}, \citenamefont {Liu}, \citenamefont {Yu},\
  and\ \citenamefont {Chen}}]{dehdashti2015realization}%
  \BibitemOpen
  \bibfield  {author} {\bibinfo {author} {\bibfnamefont {S.}~\bibnamefont
  {Dehdashti}}, \bibinfo {author} {\bibfnamefont {R.}~\bibnamefont {Li}},
  \bibinfo {author} {\bibfnamefont {J.}~\bibnamefont {Liu}}, \bibinfo {author}
  {\bibfnamefont {F.}~\bibnamefont {Yu}},\ and\ \bibinfo {author}
  {\bibfnamefont {H.}~\bibnamefont {Chen}},\ }\bibfield  {title} {\bibinfo
  {title} {Realization of non-linear coherent states by photonic lattices},\
  }\href@noop {} {\bibfield  {journal} {\bibinfo  {journal} {AIP Advances}\
  }\textbf {\bibinfo {volume} {5}},\ \bibinfo {pages} {067165} (\bibinfo {year}
  {2015})}\BibitemShut {NoStop}%
\bibitem [{\citenamefont {Castillo-Celeita}\ \emph {et~al.}(2019)\citenamefont
  {Castillo-Celeita}, \citenamefont {Bautista},\ and\ \citenamefont
  {Fernandez}}]{castillo2019polynomial}%
  \BibitemOpen
  \bibfield  {author} {\bibinfo {author} {\bibfnamefont {M.}~\bibnamefont
  {Castillo-Celeita}}, \bibinfo {author} {\bibfnamefont {E.~D.}\ \bibnamefont
  {Bautista}},\ and\ \bibinfo {author} {\bibfnamefont {D.~J.}\ \bibnamefont
  {Fernandez}},\ }\bibfield  {title} {\bibinfo {title} {Polynomial heisenberg
  algebras, multiphoton coherent states and geometric phases},\ }\href@noop {}
  {\bibfield  {journal} {\bibinfo  {journal} {Physica Scripta}\ } (\bibinfo
  {year} {2019})}\BibitemShut {NoStop}%
\bibitem [{\citenamefont {Bengtsson}\ and\ \citenamefont
  {{\.Z}yczkowski}(2017)}]{bengtsson2017geometry}%
  \BibitemOpen
  \bibfield  {author} {\bibinfo {author} {\bibfnamefont {I.}~\bibnamefont
  {Bengtsson}}\ and\ \bibinfo {author} {\bibfnamefont {K.}~\bibnamefont
  {{\.Z}yczkowski}},\ }\href@noop {} {\emph {\bibinfo {title} {Geometry of
  quantum states: an introduction to quantum entanglement}}}\ (\bibinfo
  {publisher} {Cambridge university press},\ \bibinfo {year}
  {2017})\BibitemShut {NoStop}%
\bibitem [{\citenamefont {Cari{\~n}ena}\ \emph {et~al.}(2005)\citenamefont
  {Cari{\~n}ena}, \citenamefont {Ra{\~n}ada},\ and\ \citenamefont
  {Santander}}]{carinena2005central}%
  \BibitemOpen
  \bibfield  {author} {\bibinfo {author} {\bibfnamefont {J.~F.}\ \bibnamefont
  {Cari{\~n}ena}}, \bibinfo {author} {\bibfnamefont {M.~F.}\ \bibnamefont
  {Ra{\~n}ada}},\ and\ \bibinfo {author} {\bibfnamefont {M.}~\bibnamefont
  {Santander}},\ }\bibfield  {title} {\bibinfo {title} {Central potentials on
  spaces of constant curvature: The kepler problem on the two-dimensional
  sphere s 2 and the hyperbolic plane h 2},\ }\href@noop {} {\bibfield
  {journal} {\bibinfo  {journal} {Journal of Mathematical Physics}\ }\textbf
  {\bibinfo {volume} {46}},\ \bibinfo {pages} {052702} (\bibinfo {year}
  {2005})}\BibitemShut {NoStop}%
\bibitem [{\citenamefont {Sch\"{o}lkopf}\ \emph {et~al.}(1995)\citenamefont
  {Sch\"{o}lkopf}, \citenamefont {Burgest},\ and\ \citenamefont
  {Vapnik}}]{schiilkop1995extracting}%
  \BibitemOpen
  \bibfield  {author} {\bibinfo {author} {\bibfnamefont {P.}~\bibnamefont
  {Sch\"{o}lkopf}}, \bibinfo {author} {\bibfnamefont {C.}~\bibnamefont
  {Burgest}},\ and\ \bibinfo {author} {\bibfnamefont {V.}~\bibnamefont
  {Vapnik}},\ }\bibfield  {title} {\bibinfo {title} {Extracting support data
  for a given task},\ }in\ \href@noop {} {\emph {\bibinfo {booktitle}
  {Proceedings of the 1st international conference on knowledge discovery \&
  data mining}}}\ (\bibinfo {year} {1995})\ pp.\ \bibinfo {pages}
  {252--257}\BibitemShut {NoStop}%
\bibitem [{\citenamefont {Claesen}\ \emph {et~al.}(2014)\citenamefont
  {Claesen}, \citenamefont {Simm}, \citenamefont {Popovic}, \citenamefont
  {Moreau},\ and\ \citenamefont {De~Moor}}]{claesen2014easy}%
  \BibitemOpen
  \bibfield  {author} {\bibinfo {author} {\bibfnamefont {M.}~\bibnamefont
  {Claesen}}, \bibinfo {author} {\bibfnamefont {J.}~\bibnamefont {Simm}},
  \bibinfo {author} {\bibfnamefont {D.}~\bibnamefont {Popovic}}, \bibinfo
  {author} {\bibfnamefont {Y.}~\bibnamefont {Moreau}},\ and\ \bibinfo {author}
  {\bibfnamefont {B.}~\bibnamefont {De~Moor}},\ }\bibfield  {title} {\bibinfo
  {title} {Easy hyperparameter search using optunity},\ }\href@noop {}
  {\bibfield  {journal} {\bibinfo  {journal} {arXiv preprint arXiv:1412.1114}\
  } (\bibinfo {year} {2014})}\BibitemShut {NoStop}%
\bibitem [{\citenamefont {Pedregosa}\ \emph {et~al.}(2011)\citenamefont
  {Pedregosa}, \citenamefont {Varoquaux}, \citenamefont {Gramfort},
  \citenamefont {Michel}, \citenamefont {Thirion}, \citenamefont {Grisel},
  \citenamefont {Blondel}, \citenamefont {Prettenhofer}, \citenamefont {Weiss},
  \citenamefont {Dubourg}, \citenamefont {Vanderplas}, \citenamefont {Passos},
  \citenamefont {Cournapeau}, \citenamefont {Brucher}, \citenamefont {Perrot},\
  and\ \citenamefont {Duchesnay}}]{scikit-learn}%
  \BibitemOpen
  \bibfield  {author} {\bibinfo {author} {\bibfnamefont {F.}~\bibnamefont
  {Pedregosa}}, \bibinfo {author} {\bibfnamefont {G.}~\bibnamefont
  {Varoquaux}}, \bibinfo {author} {\bibfnamefont {A.}~\bibnamefont {Gramfort}},
  \bibinfo {author} {\bibfnamefont {V.}~\bibnamefont {Michel}}, \bibinfo
  {author} {\bibfnamefont {B.}~\bibnamefont {Thirion}}, \bibinfo {author}
  {\bibfnamefont {O.}~\bibnamefont {Grisel}}, \bibinfo {author} {\bibfnamefont
  {M.}~\bibnamefont {Blondel}}, \bibinfo {author} {\bibfnamefont
  {P.}~\bibnamefont {Prettenhofer}}, \bibinfo {author} {\bibfnamefont
  {R.}~\bibnamefont {Weiss}}, \bibinfo {author} {\bibfnamefont
  {V.}~\bibnamefont {Dubourg}}, \bibinfo {author} {\bibfnamefont
  {J.}~\bibnamefont {Vanderplas}}, \bibinfo {author} {\bibfnamefont
  {A.}~\bibnamefont {Passos}}, \bibinfo {author} {\bibfnamefont
  {D.}~\bibnamefont {Cournapeau}}, \bibinfo {author} {\bibfnamefont
  {M.}~\bibnamefont {Brucher}}, \bibinfo {author} {\bibfnamefont
  {M.}~\bibnamefont {Perrot}},\ and\ \bibinfo {author} {\bibfnamefont
  {E.}~\bibnamefont {Duchesnay}},\ }\bibfield  {title} {\bibinfo {title}
  {Scikit-learn: Machine learning in {P}ython},\ }\href@noop {} {\bibfield
  {journal} {\bibinfo  {journal} {Journal of Machine Learning Research}\
  }\textbf {\bibinfo {volume} {12}},\ \bibinfo {pages} {2825} (\bibinfo {year}
  {2011})}\BibitemShut {NoStop}%
\bibitem [{\citenamefont {Stone}(1974)}]{stone1974cross}%
  \BibitemOpen
  \bibfield  {author} {\bibinfo {author} {\bibfnamefont {M.}~\bibnamefont
  {Stone}},\ }\bibfield  {title} {\bibinfo {title} {Cross-validatory choice and
  assessment of statistical predictions},\ }\href@noop {} {\bibfield  {journal}
  {\bibinfo  {journal} {Journal of the Royal Statistical Society: Series B
  (Methodological)}\ }\textbf {\bibinfo {volume} {36}},\ \bibinfo {pages} {111}
  (\bibinfo {year} {1974})}\BibitemShut {NoStop}%
\bibitem [{\citenamefont {Brandl}(2019)}]{Charles2013}%
  \BibitemOpen
  \bibfield  {author} {\bibinfo {author} {\bibfnamefont {G.~e.~a.}\
  \bibnamefont {Brandl}},\ }\href@noop {} {\bibinfo {title} {Python object
  serialization}},\ \bibinfo {howpublished}
  {\url{https://github.com/python/cpython/blob/3.7/Doc/library/pickle.rst}}
  (\bibinfo {year} {2019})\BibitemShut {NoStop}%
\bibitem [{\citenamefont {Services}(2019)}]{daly_daly_2019}%
  \BibitemOpen
  \bibfield  {author} {\bibinfo {author} {\bibfnamefont {A.~W.}\ \bibnamefont
  {Services}},\ }\href@noop {} {\bibinfo {title} {Amazon web services
  elastic-compute 2}},\ \bibinfo {howpublished}
  {https://docs.aws.amazon.com/ec2/index.html} (\bibinfo {year}
  {2019})\BibitemShut {NoStop}%
\bibitem [{\citenamefont {Osuna}\ and\ \citenamefont
  {Girosi}(1998)}]{osuna1998reducing}%
  \BibitemOpen
  \bibfield  {author} {\bibinfo {author} {\bibfnamefont {E.}~\bibnamefont
  {Osuna}}\ and\ \bibinfo {author} {\bibfnamefont {F.}~\bibnamefont {Girosi}},\
  }\bibfield  {title} {\bibinfo {title} {Reducing the run-time complexity of
  support vector machines},\ }in\ \href@noop {} {\emph {\bibinfo {booktitle}
  {International Conference on Pattern Recognition (submitted)}}}\ (\bibinfo
  {year} {1998})\BibitemShut {NoStop}%
\bibitem [{\citenamefont {Gretton}\ \emph {et~al.}(2001)\citenamefont
  {Gretton}, \citenamefont {Doucet}, \citenamefont {Herbrich}, \citenamefont
  {Rayner},\ and\ \citenamefont {Scholkopf}}]{gretton2001support}%
  \BibitemOpen
  \bibfield  {author} {\bibinfo {author} {\bibfnamefont {A.}~\bibnamefont
  {Gretton}}, \bibinfo {author} {\bibfnamefont {A.}~\bibnamefont {Doucet}},
  \bibinfo {author} {\bibfnamefont {R.}~\bibnamefont {Herbrich}}, \bibinfo
  {author} {\bibfnamefont {P.~J.}\ \bibnamefont {Rayner}},\ and\ \bibinfo
  {author} {\bibfnamefont {B.}~\bibnamefont {Scholkopf}},\ }\bibfield  {title}
  {\bibinfo {title} {Support vector regression for black-box system
  identification},\ }in\ \href@noop {} {\emph {\bibinfo {booktitle}
  {Proceedings of the 11th IEEE Signal Processing Workshop on Statistical
  Signal Processing (Cat. No. 01TH8563)}}}\ (\bibinfo {organization} {IEEE},\
  \bibinfo {year} {2001})\ pp.\ \bibinfo {pages} {341--344}\BibitemShut
  {NoStop}%
\bibitem [{\citenamefont {Faulhuber}\ \emph {et~al.}(2018)\citenamefont
  {Faulhuber}, \citenamefont {de~Gosson},\ and\ \citenamefont
  {Rottensteiner}}]{faulhuber2018gaussian}%
  \BibitemOpen
  \bibfield  {author} {\bibinfo {author} {\bibfnamefont {M.}~\bibnamefont
  {Faulhuber}}, \bibinfo {author} {\bibfnamefont {M.~A.}\ \bibnamefont
  {de~Gosson}},\ and\ \bibinfo {author} {\bibfnamefont {D.}~\bibnamefont
  {Rottensteiner}},\ }\bibfield  {title} {\bibinfo {title} {Gaussian
  distributions and phase space weyl--heisenberg frames},\ }\href@noop {}
  {\bibfield  {journal} {\bibinfo  {journal} {Applied and Computational
  Harmonic Analysis}\ } (\bibinfo {year} {2018})}\BibitemShut {NoStop}%
\bibitem [{\citenamefont {Chapelle}\ and\ \citenamefont
  {Zien}(2005)}]{chapelle2005semi}%
  \BibitemOpen
  \bibfield  {author} {\bibinfo {author} {\bibfnamefont {O.}~\bibnamefont
  {Chapelle}}\ and\ \bibinfo {author} {\bibfnamefont {A.}~\bibnamefont
  {Zien}},\ }\bibfield  {title} {\bibinfo {title} {Semi-supervised
  classification by low density separation.},\ }in\ \href@noop {} {\emph
  {\bibinfo {booktitle} {AISTATS}}},\ Vol.\ \bibinfo {volume} {2005}\ (\bibinfo
  {organization} {Citeseer},\ \bibinfo {year} {2005})\ pp.\ \bibinfo {pages}
  {57--64}\BibitemShut {NoStop}%
\bibitem [{\citenamefont {Rebentrost}\ \emph {et~al.}(2014)\citenamefont
  {Rebentrost}, \citenamefont {Mohseni},\ and\ \citenamefont
  {Lloyd}}]{rebentrost2014quantum}%
  \BibitemOpen
  \bibfield  {author} {\bibinfo {author} {\bibfnamefont {P.}~\bibnamefont
  {Rebentrost}}, \bibinfo {author} {\bibfnamefont {M.}~\bibnamefont
  {Mohseni}},\ and\ \bibinfo {author} {\bibfnamefont {S.}~\bibnamefont
  {Lloyd}},\ }\bibfield  {title} {\bibinfo {title} {Quantum support vector
  machine for big data classification},\ }\href@noop {} {\bibfield  {journal}
  {\bibinfo  {journal} {Physical review letters}\ }\textbf {\bibinfo {volume}
  {113}},\ \bibinfo {pages} {130503} (\bibinfo {year} {2014})}\BibitemShut
  {NoStop}%
\bibitem [{\citenamefont {Pitowsky}(1994)}]{pitowsky1994george}%
  \BibitemOpen
  \bibfield  {author} {\bibinfo {author} {\bibfnamefont {I.}~\bibnamefont
  {Pitowsky}},\ }\bibfield  {title} {\bibinfo {title} {George boole's
  ‘conditions of possible experience’and the quantum puzzle},\ }\href@noop
  {} {\bibfield  {journal} {\bibinfo  {journal} {The British Journal for the
  Philosophy of Science}\ }\textbf {\bibinfo {volume} {45}},\ \bibinfo {pages}
  {95} (\bibinfo {year} {1994})}\BibitemShut {NoStop}%
\bibitem [{\citenamefont {Goh}\ \emph {et~al.}(2018)\citenamefont {Goh},
  \citenamefont {Kaniewski}, \citenamefont {Wolfe}, \citenamefont
  {V{\'e}rtesi}, \citenamefont {Wu}, \citenamefont {Cai}, \citenamefont
  {Liang},\ and\ \citenamefont {Scarani}}]{goh2018geometry}%
  \BibitemOpen
  \bibfield  {author} {\bibinfo {author} {\bibfnamefont {K.~T.}\ \bibnamefont
  {Goh}}, \bibinfo {author} {\bibfnamefont {J.}~\bibnamefont {Kaniewski}},
  \bibinfo {author} {\bibfnamefont {E.}~\bibnamefont {Wolfe}}, \bibinfo
  {author} {\bibfnamefont {T.}~\bibnamefont {V{\'e}rtesi}}, \bibinfo {author}
  {\bibfnamefont {X.}~\bibnamefont {Wu}}, \bibinfo {author} {\bibfnamefont
  {Y.}~\bibnamefont {Cai}}, \bibinfo {author} {\bibfnamefont {Y.-C.}\
  \bibnamefont {Liang}},\ and\ \bibinfo {author} {\bibfnamefont
  {V.}~\bibnamefont {Scarani}},\ }\bibfield  {title} {\bibinfo {title}
  {Geometry of the set of quantum correlations},\ }\href@noop {} {\bibfield
  {journal} {\bibinfo  {journal} {Physical Review A}\ }\textbf {\bibinfo
  {volume} {97}},\ \bibinfo {pages} {022104} (\bibinfo {year}
  {2018})}\BibitemShut {NoStop}%
\bibitem [{\citenamefont {Vourdas}(2019)}]{vourdas2019probabilistic}%
  \BibitemOpen
  \bibfield  {author} {\bibinfo {author} {\bibfnamefont {A.}~\bibnamefont
  {Vourdas}},\ }\bibfield  {title} {\bibinfo {title} {Probabilistic
  inequalities and measurements in bipartite systems},\ }\href@noop {}
  {\bibfield  {journal} {\bibinfo  {journal} {Journal of Physics A:
  Mathematical and Theoretical}\ } (\bibinfo {year} {2019})}\BibitemShut
  {NoStop}%
\bibitem [{\citenamefont {Steinbrecher}\ \emph {et~al.}(2019)\citenamefont
  {Steinbrecher}, \citenamefont {Olson}, \citenamefont {Englund},\ and\
  \citenamefont {Carolan}}]{steinbrecher2019quantum}%
  \BibitemOpen
  \bibfield  {author} {\bibinfo {author} {\bibfnamefont {G.~R.}\ \bibnamefont
  {Steinbrecher}}, \bibinfo {author} {\bibfnamefont {J.~P.}\ \bibnamefont
  {Olson}}, \bibinfo {author} {\bibfnamefont {D.}~\bibnamefont {Englund}},\
  and\ \bibinfo {author} {\bibfnamefont {J.}~\bibnamefont {Carolan}},\
  }\bibfield  {title} {\bibinfo {title} {Quantum optical neural networks},\
  }\href@noop {} {\bibfield  {journal} {\bibinfo  {journal} {npj Quantum
  Information}\ }\textbf {\bibinfo {volume} {5}},\ \bibinfo {pages} {60}
  (\bibinfo {year} {2019})}\BibitemShut {NoStop}%
\end{thebibliography}%


\begin{thebibliography}{10}

\bibitem{boser1992training}
Bernhard~E Boser, Isabelle~M Guyon, and Vladimir~N Vapnik.
\newblock A training algorithm for optimal margin classifiers.
\newblock In {\em Proceedings of the fifth annual workshop on Computational
  learning theory}, pages 144--152. ACM, 1992.

\bibitem{cortes1995support}
Corinna Cortes and Vladimir Vapnik.
\newblock Support-vector networks.
\newblock {\em Machine learning}, 20(3):273--297, 1995.

\bibitem{Haykin08}
Simon~O. Haykin.
\newblock {\em Neural Networks and Learning Machines: A Comprehensive
  Foundation}.
\newblock Pearson (3rd edition), 2008.

\bibitem{Bhattacharyya11}
Amit Konar D.N Tibarewala Atulya K.~Nagar Saugat~Bhattacharyya,
  Anwesha~Khasnobish.
\newblock Performance analysis of left/right hand movement classification from
  eeg signal by intelligent algorithms.
\newblock In {\em IEEE Symposium on Computational Intelligence, Cognitive
  Algorithms, Mind, and Brain (CCMB)}, pages 1--8, 2011.

\bibitem{schuld2019quantum}
Maria Schuld and Nathan Killoran.
\newblock Quantum machine learning in feature hilbert spaces.
\newblock {\em Physical review letters}, 122(4):040504, 2019.

\bibitem{schuld2019machine}
Maria Schuld.
\newblock Machine learning in quantum spaces, 2019.

\bibitem{havlivcek2019supervised}
Vojt{\v{e}}ch Havl{\'\i}{\v{c}}ek, Antonio~D C{\'o}rcoles, Kristan Temme,
  Aram~W Harrow, Abhinav Kandala, Jerry~M Chow, and Jay~M Gambetta.
\newblock Supervised learning with quantum-enhanced feature spaces.
\newblock {\em Nature}, 567(7747):209, 2019.

\bibitem{mehta2019high}
Pankaj Mehta, Marin Bukov, Ching-Hao Wang, Alexandre~GR Day, Clint Richardson,
  Charles~K Fisher, and David~J Schwab.
\newblock A high-bias, low-variance introduction to machine learning for
  physicists.
\newblock {\em Physics reports}, 2019.

\bibitem{killoran2019strawberry}
Nathan Killoran, Josh Izaac, Nicol{\'a}s Quesada, Ville Bergholm, Matthew Amy,
  and Christian Weedbrook.
\newblock Strawberry fields: A software platform for photonic quantum
  computing.
\newblock {\em Quantum}, 3:129, 2019.

\bibitem{killoran2018continuous}
Nathan Killoran, Thomas~R Bromley, Juan~Miguel Arrazola, Maria Schuld,
  Nicol{\'a}s Quesada, and Seth Lloyd.
\newblock Continuous-variable quantum neural networks.
\newblock {\em arXiv preprint arXiv:1806.06871}, 2018.

\bibitem{srinivasan2018learning}
Siddarth Srinivasan, Carlton Downey, and Byron Boots.
\newblock Learning and inference in hilbert space with quantum graphical
  models.
\newblock In {\em Advances in Neural Information Processing Systems}, pages
  10338--10347, 2018.

\bibitem{adhikary2019supervised}
Soumik Adhikary, Siddharth Dangwal, and Debanjan Bhowmik.
\newblock Supervised learning with a quantum classifier using a multi-level
  system.
\newblock {\em arXiv preprint arXiv:1908.08385}, 2019.

\bibitem{bartkiewicz2019experimental}
Karol Bartkiewicz, Clemens Gneiting, Anton{\'\i}n {\v{C}}ernoch, Kate{\v{r}}ina
  Jir{\'a}kov{\'a}, Karel Lemr, and Franco Nori.
\newblock Experimental kernel-based quantum machine learning in finite feature
  space.
\newblock {\em arXiv preprint arXiv:1906.04137}, 2019.

\bibitem{kubler2019quantum}
Jonas~M K{\"u}bler, Krikamol Muandet, and Bernhard Sch{\"o}lkopf.
\newblock Quantum mean embedding of probability distributions.
\newblock {\em Physical Review Research}, 1(3):033159, 2019.

\bibitem{combescure2012coherent}
Monique Combescure and Didier Robert.
\newblock {\em Coherent states and applications in mathematical physics}.
\newblock Springer Science \& Business Media, 2012.

\bibitem{ali2000coherent}
Syed~Twareque Ali, Jean-Pierre Antoine, Jean-Pierre Gazeau, et~al.
\newblock {\em Coherent states, wavelets and their generalizations}, volume~3.
\newblock Springer, 2000.

\bibitem{Ali06}
Shawkat Ali and Kate A.Smith-Miles.
\newblock A meta-learning approach to automatic kernel selection for support
  vector machines.
\newblock {\em Neurocomputing}, 70:173--186, 2006.

\bibitem{dehdashti2013coherent}
Sh~Dehdashti, A~Mahdifar, and R~Roknizadeh.
\newblock Coherent state of $\alpha$-deformed weyl--heisenberg algebra.
\newblock {\em International Journal of Geometric Methods in Modern Physics},
  10(05):1350014, 2013.

\bibitem{dehdashti2015realization}
Shahram Dehdashti, Rujiang Li, Jiarui Liu, Faxin Yu, and Hongsheng Chen.
\newblock Realization of non-linear coherent states by photonic lattices.
\newblock {\em AIP Advances}, 5(6):067165, 2015.

\bibitem{castillo2019polynomial}
Miguel Castillo-Celeita, Erik~Diaz Bautista, and David~J Fernandez.
\newblock Polynomial heisenberg algebras, multiphoton coherent states and
  geometric phases.
\newblock {\em Physica Scripta}, 2019.

\bibitem{bengtsson2017geometry}
Ingemar Bengtsson and Karol {\.Z}yczkowski.
\newblock {\em Geometry of quantum states: an introduction to quantum
  entanglement}.
\newblock Cambridge university press, 2017.

\bibitem{carinena2005central}
Jos{\'e}~F Cari{\~n}ena, Manuel~F Ra{\~n}ada, and Mariano Santander.
\newblock Central potentials on spaces of constant curvature: The kepler
  problem on the two-dimensional sphere s 2 and the hyperbolic plane h 2.
\newblock {\em Journal of Mathematical Physics}, 46(5):052702, 2005.

\end{thebibliography}

\end{document}